# Does Structured Intent Representation Generalize?

## A Cross-Language, Cross-Model Empirical Study of 5W3H Prompting

**PENG Gang**

Huizhou Lateni AI Technology Co., Ltd., Huizhou, China; Huizhou University, Huizhou, China penggangjp@gmail.com

**Abstract**

Unstructured natural language prompts suffer from *intent transmission loss* — the gap between what a user intends and what an AI model produces. Our prior work introduced PPS (Prompt Protocol Specification), a 5W3H-based structured intent framework, and demonstrated its effectiveness in Chinese. This paper extends that work along three dimensions: (1) two additional languages (English and Japanese), (2) a fourth experimental condition in which a user's simple prompt is automatically expanded into a full 5W3H specification via an AI-assisted authoring interface (Condition D), and (3) a new research question on cross-model output consistency as a measure of protocol-level reliability.

Across 2,160 model outputs (3 languages × 4 conditions × 3 LLMs × 60 tasks), we find that: (a) AI-expanded 5W3H prompts (Condition D) show no statistically significant difference in goal alignment from manually crafted 5W3H prompts (Condition C) across all three languages, while requiring only a single-sentence input from the user; (b) structured PPS conditions often reduce or reshape cross-model output variance, though this effect is not uniform across all languages and metrics — the strongest evidence comes from the reduction of spurious low variance in unconstrained baselines; and (c) unstructured prompts (Condition A) exhibit a systematic dual-inflation bias — artificially high composite scores and artificially low cross-model variance — that challenges their validity as an evaluation baseline. These findings suggest that structured 5W3H representations can improve intent alignment and accessibility across languages and models, particularly when AI-assisted authoring lowers the barrier for non-expert users.

## 1. Introduction

### 1.1 The Intent Transmission Problem

The rapid deployment of large language models (LLMs) has created a fundamental asymmetry in human-AI communication: while models possess broad generative capability, users typically express their intent through brief, unstructured natural language prompts that incompletely specify the desired output. This gap — which we term *intent transmission loss* — manifests as unpredictable output quality, inconsistency across sessions, and high dependence on users' prompt engineering skill.

Empirical studies confirm the severity of this problem. Small changes in prompt wording can produce dramatically different outputs from the same model [5, 25], and the same task description yields substantially different results across different LLMs [12, 13]. For practical applications — professional writing, research assistance, business analysis — this unpredictability is a critical barrier to adoption.

Existing approaches to this problem fall into two broad categories. *Prompt engineering* techniques (few-shot examples, chain-of-thought, role assignment) improve outputs for individual tasks but require expertise and do not generalize across tasks or models. *Structured prompting* approaches (XML templates, JSON instructions) impose format constraints but lack a principled framework for capturing the full dimensionality of user intent.

**1.2 PPS: Structured Intent Representation**

In our prior work (arXiv:2603.18976), we proposed PPS (Prompt Protocol Specification), a structured intent representation framework organized around eight dimensions drawn from the journalistic 5W3H framework: *What* (task and KPIs), *Why* (goals and constraints), *Who* (persona and audience), *When* (timing), *Where* (context and platform), *How-to-do* (method and steps), *How-much* (quantitative requirements), and *How-feel* (tone and style).

Unlike prompt engineering techniques that optimize individual prompts, PPS frames intent representation as a structured communication layer for human-AI interaction: the goal is to define a standard representation that enables more reliable intent transmission regardless of the specific model receiving it. This framing motivates the evaluation criteria used in this paper: not only output quality, but also cross-model consistency and cross-language generalizability.

Our prior work [27] established the effectiveness of PPS in Chinese across three experimental conditions (A: unstructured, B: raw JSON, C: PPS natural language rendering) and three LLMs. Two limitations remained: the study was monolingual, and condition C required users to manually specify all eight dimensions — a significant barrier for non-expert users.

**1.3 This Paper: Extension Across Languages, Models, and Authoring Methods**

This paper addresses both limitations and introduces a new evaluation dimension:

**Extension 1: Cross-language generalizability (RQ2).** We replicate the full experiment in English and Japanese, using the same 60 tasks translated and adapted for each language. If PPS's structured representation is language-agnostic in its intent-alignment effects, its effectiveness should generalize across languages.

**Extension 2: AI-assisted intent expansion (Condition D).** We introduce a fourth experimental condition in which the user's simple prompt (identical to Condition A) is automatically expanded into a full 5W3H specification by an AI-assisted authoring interface, then rendered as natural language before being sent to the target model. This condition tests whether AI-mediated structuring can match the quality of manual structuring while eliminating the expertise barrier.

**Extension 3: Cross-model consistency as a structured-layer property (RQ3).** We introduce a new evaluation metric: the variance of output scores across models for the same task and condition. A structured representation that reliably transmits intent should tend to produce more consistent outputs regardless of the receiving model; we test whether PPS-structured conditions exhibit different cross-model variance patterns compared to unstructured conditions.

### 1.4 Contributions

1. **First three-language PPS empirical study** (Chinese, English, Japanese), demonstrating cross-lingual generalizability across 2,160 model outputs.
2. **Condition D**: AI-expanded 5W3H prompts show no statistically significant difference from manually crafted PPS prompts on goal alignment, with a single-sentence user input — democratizing access to structured prompting.
3. **RQ3 (cross-model consistency)**: Structured conditions exhibit significantly different cross-model variance patterns from unstructured baselines, with evidence that Condition A's apparent consistency is an artifact of unconstrained generation.
4. **Dual-inflation analysis of Condition A**: We show that unstructured prompts systematically inflate both composite quality scores and apparent cross-model consistency, challenging their use as isolated baselines.

---

## 2. Related Work

### 2.1 Prompt Engineering

Since Brown et al. demonstrated that GPT-3 responds to in-context examples [1], prompt engineering has evolved rapidly. Ouyang et al.'s InstructGPT showed that human feedback fine-tuning dramatically improves instruction following [15]. Wei et al.'s chain-of-thought prompting demonstrated that inserting intermediate reasoning steps substantially improves complex task performance [2]. Subsequent work extended this to tree-of-thought [3] and other structured reasoning decompositions. Shanahan et al. analyzed role-play as a prompting mechanism, finding that persona assignment significantly shapes model output style [24]. Zhou et al. demonstrated that LLMs can themselves generate effective prompts when given meta-level instructions [18].

Comprehensive surveys by Liu et al. [4] and Schulhoff et al. [5] catalog dozens of prompting techniques and templates; Sahoo et al. provide a focused survey of techniques and their applications [25]. However, all these techniques address *how AI should reason*, not *how users should express intent*. Chain-of-thought instructs the AI; PPS structures the user. The two approaches are orthogonal and complementary: a PPS-formatted prompt can include CoT instructions in its `how_to_do` dimension.

### 2.2 Structured and Template-Based Prompting

Several informal frameworks have proposed structured components for prompts. Schulhoff et al.'s survey [5] catalogs frameworks including ICIO (Identity, Context, Input, Output), CO-STAR (Context, Objective, Style, Tone, Audience, Response), and CRISPE, among others. PromptSource [22] and Self-Instruct [20] have operationalized prompt templates for model training purposes. Reynolds and McDonell analyzed prompt programming as an HCI concern, distinguishing between prompts as *interfaces* versus *programs* [6].

Most directly related to our work, Liu et al. [7] conducted a CHI 2024 study of users' needs for structured LLM output, finding that users consistently identify the absence of formal specification as a core barrier — a gap PPS is designed to address. Mishra et al. demonstrated that reframing instructional prompts to match the model's expected format significantly improves instruction following [19]. ChainForge [17] provides visual tooling for prompt hypothesis testing across multiple model configurations — a complementary approach that could be used to validate PPS variants across models.

PPS differs from these frameworks primarily in offering an explicitly structured, machine-readable format with cryptographic integrity (SHA-256), persistent Instruction IDs for cross-session traceability, and a rendering layer that converts the structured representation into natural language before submission. This paper extends our prior empirical validation of PPS [27] to cross-language and cross-model settings, and introduces AI-assisted intent expansion (Condition D) as a democratization pathway.

To our knowledge, no prior work has systematically evaluated whether a structured intent representation framework generalizes across languages and reduces cross-model output variance — the two central questions of this paper.

## 2.3 Human-AI Interaction and Accessibility

From an HCI perspective, considerable work has studied how users interact with AI systems. Amershi et al.'s guidelines for human-AI interaction [8] identify eighteen principles, many of which PPS operationalizes structurally (e.g., "make clear what the system can do," "remember recent interactions"). Zamfirescu-Pereira et al. [11] conducted an empirical study of why non-expert users fail at prompt engineering, identifying lack of mental model and unpredictable AI behavior as core barriers — both addressed by PPS's structured specification. Jiang et al. [9] demonstrate that users invest significant iterative effort in prompt prototyping, motivating tool support that reduces this burden.

Salemi et al. [10] introduce LaMP, a benchmark for personalized LLMs, demonstrating that incorporating user-specific context substantially improves output relevance — a result consistent with PPS's emphasis on explicit Who and How-feel dimensions. Liao and Vaughan [21] argue that AI transparency in the LLM era requires new user-centered frameworks that make AI behavior predictable, a goal PPS addresses through structured intent specifications and field locks. Kim et al. [23] show that how models express uncertainty affects user reliance and trust; PPS's explicit constraint specifications can partially mitigate this by grounding outputs in verifiable intent.

Our work is positioned at the intersection of this literature and the goal of structured intent representation. Loosely inspired by how layered communication protocols standardize data transmission to improve reliability across heterogeneous systems [16], PPS aims to provide a reusable representation layer that makes intent transmission more reliable across different AI models and user contexts. This framing motivates our focus on consistency and generalizability rather than peak performance, and distinguishes our cross-model consistency metric (RQ3) from standard benchmarking.

## 2.4 Evaluation of LLM Outputs

Zheng et al. introduced LLM-as-Judge, demonstrating that strong LLM evaluators correlate well with human judgments on MT-Bench, while also documenting potential self-preference bias [12]. LMSYS Chatbot Arena uses large-scale pairwise human preference comparisons to benchmark models in a more ecologically valid setting [13]. Min et al.'s FActScore [14] evaluates factual precision against specific knowledge sources; our `goal_alignment` metric extends this principle to intent alignment — grounding quality assessment in the user's specified (or inferred) intent rather than generic quality standards.

Our prior work [27] introduced goal_alignment as a user-centered evaluation dimension, complementing generic quality metrics by requiring the judge to assess fit between output and the user's actual purpose. The current paper extends this evaluation framework to three languages and adds cross-model standard deviation as a protocol-level consistency metric. Cross-model consistency has received less systematic attention as an evaluation criterion, though it is implicitly addressed in benchmarking studies comparing model capabilities [13].

---

# 3. The PPS Framework and Experimental Conditions

## 3.1 5W3H Intent Dimensions

PPS organizes user intent into eight dimensions:

| Dimension | Description | Example (Travel domain) |
| --- | --- | --- |
| **What** | Core task + success criteria (KPIs) | Write a 5-day Tokyo self-guided travel guide; cover ≥5 attractions |
| **Why** | Goals and constraints | Help first-time Japan visitors; exclude closed venues |
| **Who** | Persona and target audience | Travel blogger; urban professionals aged 25–35 |
| **When** | Timing and deadline | Spring travel (March–May) |
| **Where** | Deployment context and jurisdiction | Blog platform; Japan |
| **How-to-do** | Method and step sequence | Overview → itinerary → attractions → transport → tips |
| **How-much** | Quantitative requirements | 1,200–1,500 words; include cost table |
| **How-feel** | Tone, style, and reader level | Enthusiastic; beginner-friendly |

The eight dimensions are designed to collectively specify the pragmatic, contextual, and aesthetic requirements of a task — the dimensions along which unstructured prompts most commonly underspecify.

## 3.2 Four Experimental Conditions

We evaluate four conditions representing a spectrum from unstructured to fully structured intent representation:

**Condition A — Simple Prompt (Baseline).** A short, natural language task description representative of typical user behavior. Example: *"Please write a 5-day Tokyo travel guide including major attractions, budget information, and transportation tips."* Temperature = 0.7 to simulate realistic user variability.

**Condition B — PPS Raw JSON.** The complete PPS specification in raw JSON format, wrapped in a prompt instructing the model to follow it strictly. This condition tests whether structured information alone, without rendering, improves outputs. Temperature = 0, seed = 42.

**Condition C — PPS Natural Language Rendering.** The same PPS body as Condition B, rendered into structured natural language with labeled section headers (e.g., "Task Goal (What): ...", "Purpose (Why): ..."). Headers are language-appropriate (Chinese/English/Japanese). This condition tests the hypothesis that rendering the structure into the model's native format is necessary for effective intent transmission. Temperature = 0, seed = 42.

**Condition D — AI-Expanded 5W3H.** The user's simple prompt (identical in content to Condition A) is submitted to an AI-assisted authoring interface as the *What* dimension seed. The interface uses an LLM (DeepSeek-V3, `deepseek-chat` API) to automatically expand the remaining seven dimensions, producing a ~1,000-word structured natural language prompt identical in format to Condition C. This condition tests whether AI-mediated expansion can match manually crafted PPS while requiring only a single-sentence user input. Temperature = 0, seed = 42.

**Key design principle:** Conditions A, B, C, and D share the same task core — Condition A's content equals the *What* dimension of B, C, and D. This helps reduce task-content differences across conditions, though A-versus-structured comparisons still involve additional confounds (temperature and specification richness) discussed in §6.7.

**Note on temperature settings:** Condition A uses temperature = 0.7 to simulate realistic user interaction conditions, while Conditions B, C, and D use temperature = 0 for deterministic output. This design choice prioritizes ecological validity (Condition A reflects how users actually interact with LLMs) over strict experimental control. As a consequence, observed differences between A and other conditions reflect the *joint effect* of structural completeness and sampling variability. We address this confound in §6.7 (Limitations).

**Note on D as a conservative lower bound:** In this experiment, Condition D prompts are used as generated, without user review or modification. In practice, users can inspect and refine any dimension before submission. This suggests that AI-assisted authoring may be practically more effective than the present lower-bound estimate, but confirming this requires a direct user study.

## 3.3 Language Adaptation

For each language, all 60 tasks were translated and culturally adapted, preserving cognitive difficulty and domain coverage. Condition A prompts were translated directly. PPS body fields (What, Why, Who, etc.) were translated at the field level. Section headers in Condition C rendering were language-appropriate:

- **Chinese:** 任务目标 (What) / 执行原因 (Why) / ...
- **English:** Task Goal (What) / Purpose (Why) / ...
- **Japanese:** タスク目標 (What) / 実行理由 (Why) / ...

For Condition D in Japanese, the lateni.com authoring interface was called with `language: "ja"`, producing fully Japanese-language 5W3H expansions.

---

## 4. Experimental Design

### 4.1 Task Corpus

We constructed a corpus of 60 tasks across three domains:

- **Travel** (20 tasks): Destination travel guides for cities across Asia and Europe
- **Business** (20 tasks): Market analysis reports for emerging industry sectors
- **Technical** (20 tasks): Technical explanation and research planning documents

Tasks vary in openness and audience specificity: Travel tasks tend to be more open-ended with broad audiences; Business tasks have more defined professional audiences; Technical tasks involve more constrained, domain-specific requirements. This variation allows us to examine whether structured prompting provides differential benefits across task types. We defer the discussion of ambiguity-related patterns to the Results section (§5) rather than pre-imposing a classification.

### 4.2 Models

Three LLMs were evaluated as generation models:

| Model | API model ID | Provider | Endpoint |
| --- | --- | --- | --- |
| DeepSeek-V3 | `deepseek-chat` | DeepSeek | api.deepseek.com |
| Qwen-max | `qwen-max` | Alibaba Cloud | dashscope.aliyuncs.com |
| Kimi | `moonshot-v1-8k` | Moonshot AI | api.moonshot.cn |

These models were selected for their strong multilingual capabilities and accessibility. All three support Chinese, English, and Japanese. The API model ID `deepseek-chat` refers to DeepSeek-V3, consistent with the model used in our prior work [27]. A separate DeepSeek-V3 instance (same API endpoint) served as the LLM-as-Judge.

### 4.3 Evaluation Metrics

**Composite Score (LLM-as-Judge, 1–5 per dimension):** Five dimensions scored independently, averaged to produce a composite: - *task_completion*: Did the output fully accomplish the stated task? - *structure*: Is the output logically organized? - *specificity*: Are claims concrete and evidence-based? - *constraint_adherence*: Were explicit constraints followed? - *overall_quality*: Holistic domain-expert assessment

**Goal Alignment (GA, 1–5):** An independent judge score assessing how well the output matches the user's *actual intent*, not generic quality. For Condition A (no explicit constraints), the judge infers the most likely user intent from context. For Conditions C and D, the judge verifies alignment with the explicit 5W3H specification. This asymmetry captures the "free pass" effect in Condition A scoring.

**Cross-Model Standard Deviation (σ):** For each task × condition pair, we compute the standard deviation of scores across the three models. Lower σ indicates more consistent outputs — a desirable protocol property.

### 4.4 Statistical Analysis

We used Mann-Whitney U tests (two-tailed) for all pairwise comparisons between conditions, with Cohen's *d* as effect size measure. For cross-model variance comparisons across conditions, we used the Kruskal-Wallis H test. Statistical significance threshold: $p < 0.05$.

### 4.5 Data Summary

| Language | Conditions | Models | Tasks | Total Outputs |
| --- | --- | --- | --- | --- |
| Chinese | A, B, C, D | 3 | 60 | 720 |
| English | A, B, C, D | 3 | 60 | 720 |
| Japanese | A, B, C, D | 3 | 60 | 720 |
| **Total** | | | | **2,160** |

## 5. Results

### 5.1 RQ1: Does PPS Improve Output Quality and Goal Alignment?

Table 1 presents mean composite scores and goal alignment scores for all four conditions across three languages.

**Table 1. Mean Composite Score and Goal Alignment (mean ± SD, N=180 per cell; bold = highest value per column)**

| Condition | ZH Composite | ZH GA | EN Composite | EN GA | JA Composite | JA GA |
|---|---|---|---|---|---|---|
| A: Simple | 4.838 ± 0.174 | 4.344 ± 0.828 | 4.126 ± 1.139 | 4.489 ± 0.787 | 4.699 ± 0.325 | 4.639 ± 0.567 |
| B: Raw JSON | 4.000 ± 0.741 | 4.094 ± 0.857 | 2.322 ± 0.539 | 2.450 ± 0.953 | 3.169 ± 0.833 | 3.394 ± 1.038 |
| C: PPS NL | 4.527 ± 0.508 | **4.606 ± 0.544** | 2.899 ± 1.370 | 4.228 ± 0.700 | 4.170 ± 0.855 | 4.439 ± 0.833 |
| D: AI-Expanded | 4.324 ± 0.753 | 4.494 ± 0.815 | **3.482 ± 0.681** | 4.233 ± 0.792 | 4.071 ± 0.809 | 4.389 ± 0.861 |

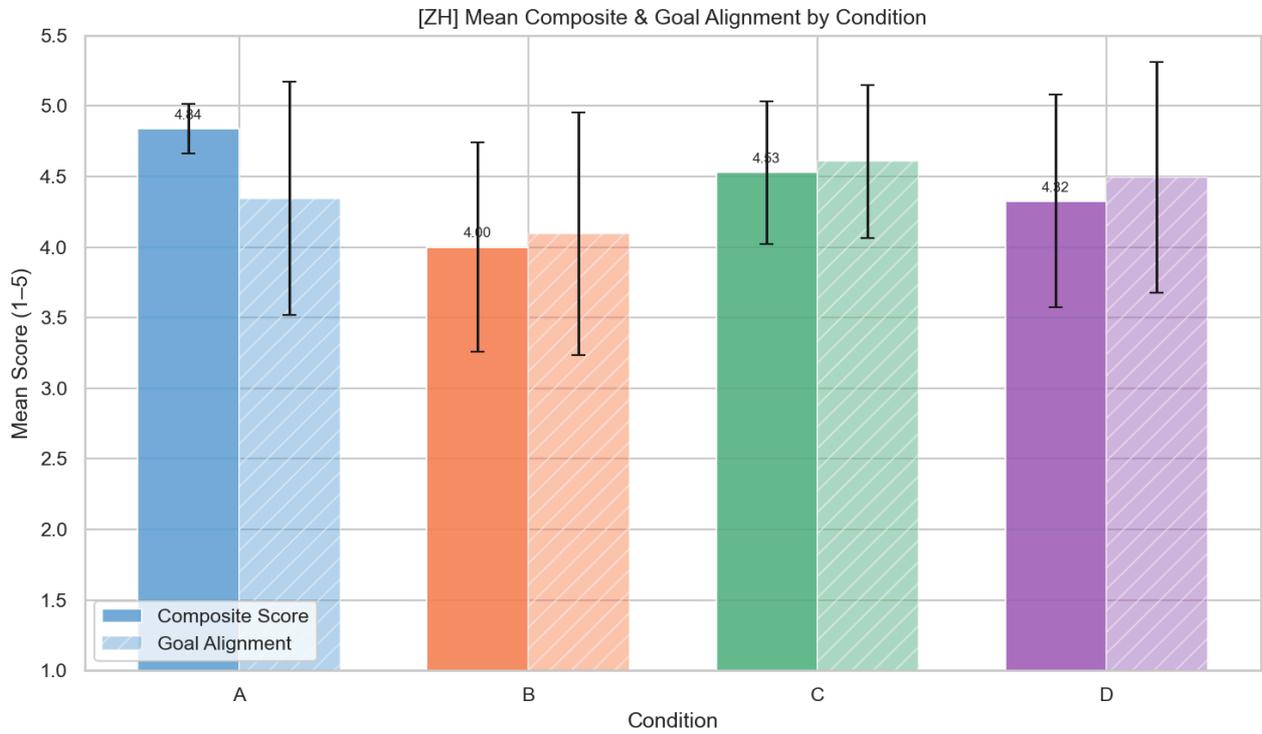

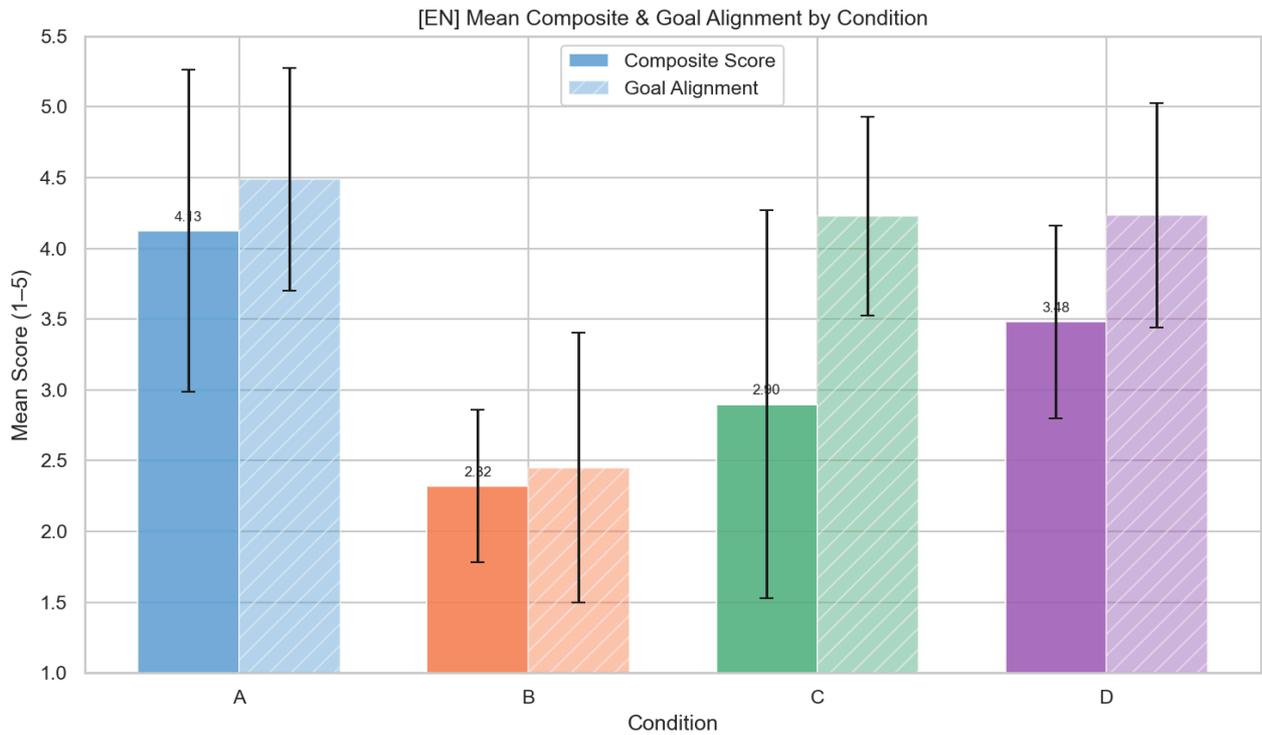

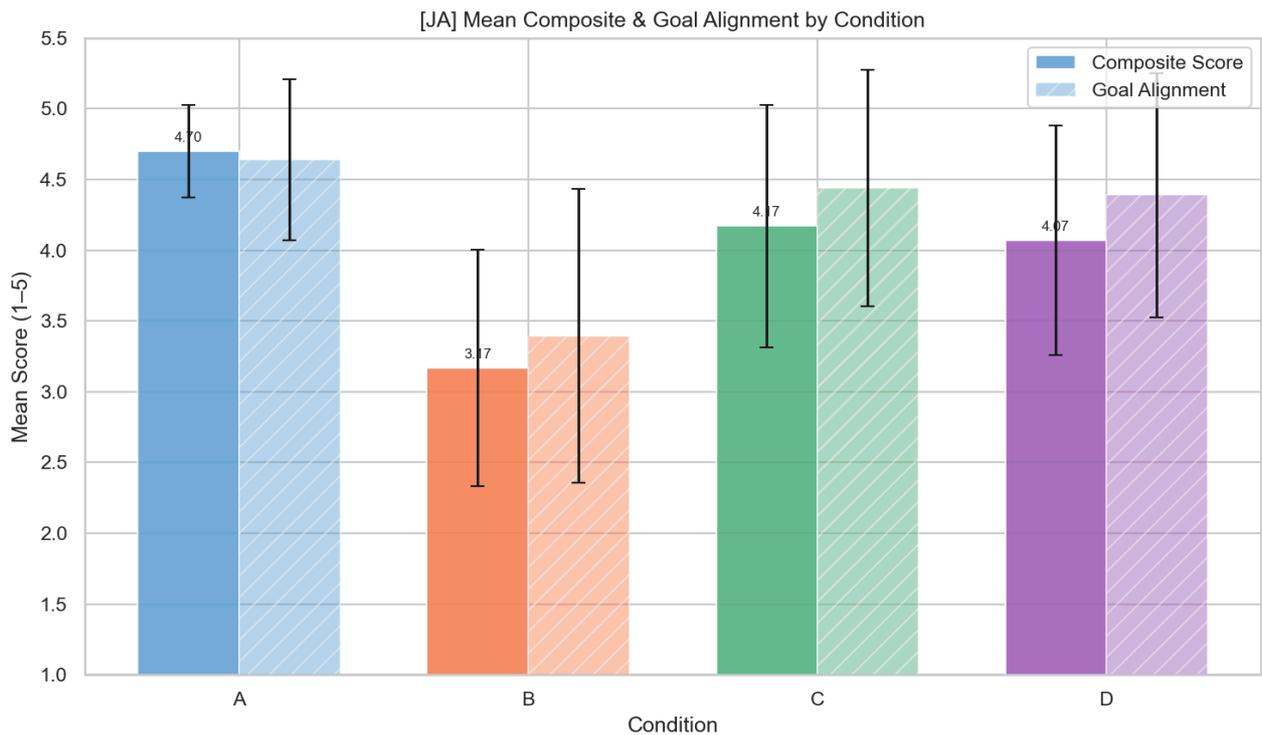

*Figure 1 (a–c). Mean composite score and goal alignment by condition across three languages. Error bars indicate ±1 SD. Exact values are reported in Table 1. Note: Condition A's high composite scores reflect baseline inflation (§6.2); GA is the more meaningful comparison.*

**Finding 1.1: Raw JSON (B) fails across all languages.** Condition B produced the lowest scores in all three languages on both metrics, with particularly severe degradation in English (composite: 2.322, GA: 2.450). This confirms that structured information alone, without appropriate rendering, is insufficient — and in English is actively harmful, likely because English-language LLMs are more sensitive to format-task mismatches in JSON-wrapped prompts.

**Finding 1.2: C and D significantly outperform B on goal alignment (all languages).** Mann-Whitney tests confirm C > B and D > B on GA across all three languages (all $p < 0.001$, Cohen's $d > 1.0$ in most comparisons; Table 2). This establishes that the rendering layer is essential: the same 5W3H information, properly rendered, produces dramatically better alignment with user intent than raw JSON.

**Finding 1.3: The "A baseline inflation" effect.** Condition A achieves the highest composite scores in all three languages. However, this should be interpreted with caution. Composite scores reward *constraint adherence* — but Condition A specifies no explicit constraints, so the judge cannot penalize for constraint violations. This creates a ceiling effect on composite scoring that makes A appear artificially strong on this metric.

The goal alignment metric partially corrects for this: in Chinese, C outperforms A on GA (4.606 vs. 4.344, $p = 0.013$), indicating that despite A's high composite score, C better satisfies the user's actual intent. Notably, D vs. A on GA in Chinese approaches but does not reach significance ($p = 0.055$; Table B1), suggesting D's advantage over A is weaker than C's — consistent with D being a conservative lower bound. In English, the direction reverses: A achieves higher GA than C (4.489 vs. 4.228, $p < 0.001$), a cross-language divergence we discuss in §6.4. In Japanese, A's GA also remains high (4.639), reflecting strong baseline model performance on open-ended tasks. We discuss the dual-inflation phenomenon in detail in §5.3 and §6.2.

**Table 2. Key Pairwise Comparisons — Composite Score and Goal Alignment**

| Language | Comparison | Metric | Δ | *p* | Cohen's *d* |
| --- | --- | --- | --- | --- | --- |
| ZH | C vs. B | Composite | +0.527 | <0.001 | +0.829 |
| ZH | C vs. B | GA | +0.511 | <0.001 | +0.712 |
| ZH | D vs. B | Composite | +0.324 | <0.001 | +0.434 |
| ZH | D vs. B | GA | +0.400 | <0.001 | +0.478 |
| ZH | C vs. A | GA | +0.261 | 0.013 | +0.373 |
| EN | C vs. B | GA | +1.778 | <0.001 | +2.127 |
| EN | D vs. B | GA | +1.783 | <0.001 | +2.036 |
| EN | D vs. C | Composite | +0.583 | <0.001 | +0.539 |
| JA | C vs. B | Composite | +1.001 | <0.001 | +1.186 |
| JA | C vs. B | GA | +1.044 | <0.001 | +1.109 |
| JA | D vs. B | Composite | +0.902 | <0.001 | +1.099 |
| JA | D vs. B | GA | +0.994 | <0.001 | +1.042 |

## 5.2 RQ2: Does PPS Generalize Across Languages?

We assess cross-language generalizability by examining whether the ordinal relationship among conditions is consistent across Chinese, English, and Japanese.

**Finding 2.1: Four consistent patterns hold across all three languages.**

The following relationships hold in all three languages (all statistically significant at $p < 0.05$ unless noted):

1. **C > B** on both composite and GA: Rendering is necessary across all languages
2. **D > B** on both composite and GA: AI-expanded prompts consistently outperform raw JSON
3. **No significant difference between C and D**: AI expansion matches manual crafting on goal alignment (see §5.2.1)
4. **B ≪ A** on composite: Raw JSON is not an effective delivery mechanism in any language

The consistency of these patterns across three typologically distinct languages — Chinese (tonal, character-based), English (Indo-European), and Japanese (agglutinative, mixed scripts) — provides strong evidence for the language-agnostic nature of PPS's effectiveness.

**Finding 2.2: Magnitude differences reflect language-model capability baselines.**

While the ordinal patterns are consistent, effect magnitudes differ. The composite score gap between B and A is largest in English ($\Delta = -1.803$) and smallest in Chinese ($\Delta = -0.838$). This likely reflects that English-language models have stronger general instruction-following capabilities, making the format disruption of raw JSON relatively more costly when mixed with content instructions.

In Chinese, A's composite advantage over C is larger than in English/Japanese (+0.311 vs. +1.227 and +0.529). This is consistent with the finding in [27] that Chinese-language PPS prompts sometimes over-constrain models with strong domain knowledge, slightly reducing composite scores while improving goal alignment.

### 5.2.1 The Central Finding: No Significant Difference Between C and D

The most practically significant finding of this paper is that no statistically significant difference was observed between Conditions C and D on goal alignment in any of the three languages:

| Language | C vs. D (GA) | Δ | $p$ | Cohen's $d$ |
|---|---|---|---|---|
| Chinese | C ≈ D | +0.111 | 0.812 | −0.160 |
| English | C ≈ D | +0.006 | 0.633 | +0.007 |
| Japanese | C ≈ D | +0.050 | 0.772 | −0.059 |

In no language does the difference between manually crafted PPS (C) and AI-expanded PPS (D) reach statistical significance on goal alignment. This means that:

> *A user who provides a single-sentence task description achieves a comparable level of intent alignment to a user who manually specifies all eight 5W3H dimensions — with no statistically significant difference observed — provided the single-sentence input is expanded by an AI-assisted authoring interface.*

Given that Condition D uses AI-generated expansions without user modification, it represents a conservative lower bound. Users who review and adjust the AI-generated dimensions may achieve even better results, though this requires confirmation through a direct user study.

---

### 5.3 RQ3: Does PPS Reduce Cross-Model Output Variance?

A structured representation that reliably transmits intent should tend to produce consistent results regardless of the receiving model. We operationalize this as cross-model consistency: for each task × condition, we compute the standard deviation of scores across the three models (σ), then average across tasks.

**Table 3. Cross-Model Standard Deviation by Condition and Language** *(Lower σ = more consistent outputs across models; bold = lowest σ per column)*

| Condition | ZH σ (composite) | EN σ (composite) | JA σ (composite) |
|---|---|---|---|
| A | **0.160** | 0.778 | **0.290** |
| B | 0.743 | 0.464 | 0.792 |
| C | 0.452 | 0.806 | 0.802 |
| D | 0.571 | **0.587** | **0.622** |

Kruskal-Wallis tests confirm significant differences in cross-model variance across conditions in all three languages: - Chinese: H = 104.689, $p < 0.001$ - English: H = 8.545, $p = 0.036$ - Japanese: H = 67.771, $p < 0.001$

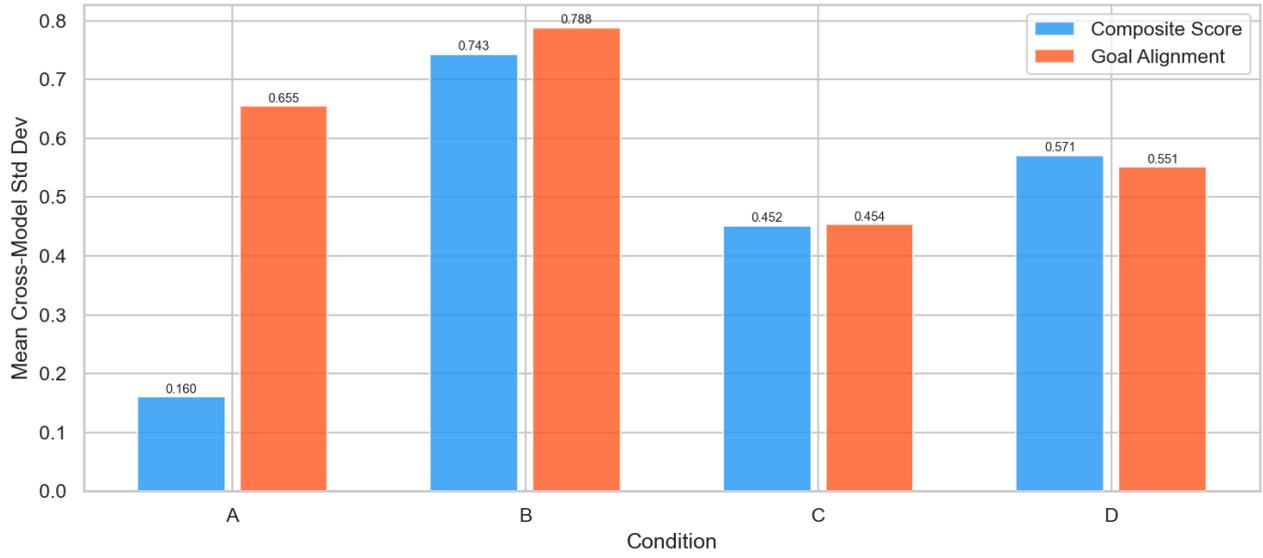

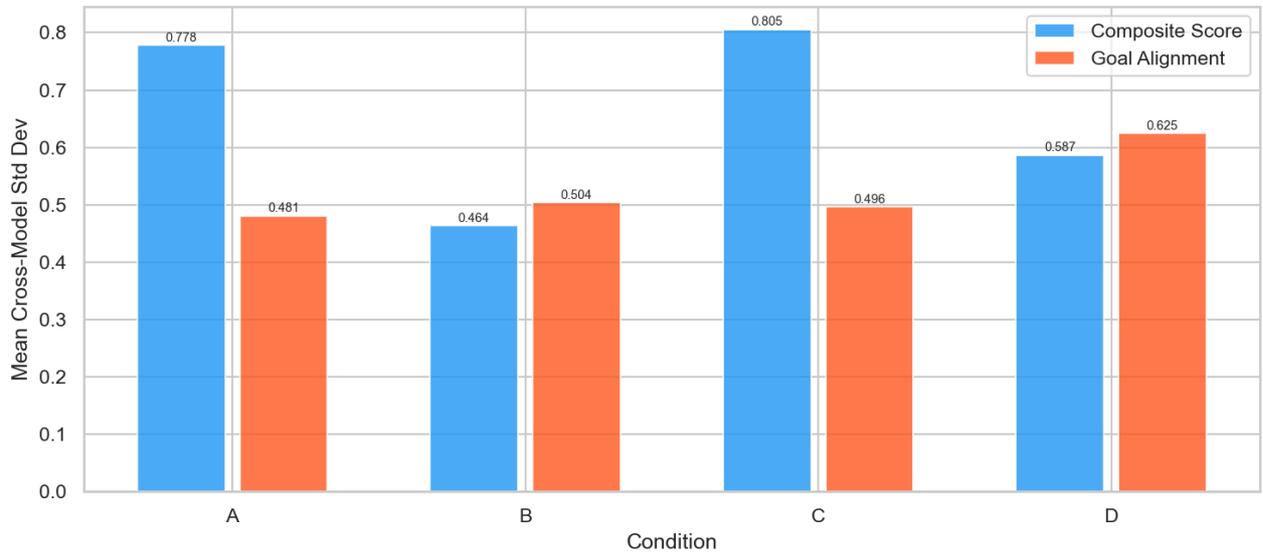

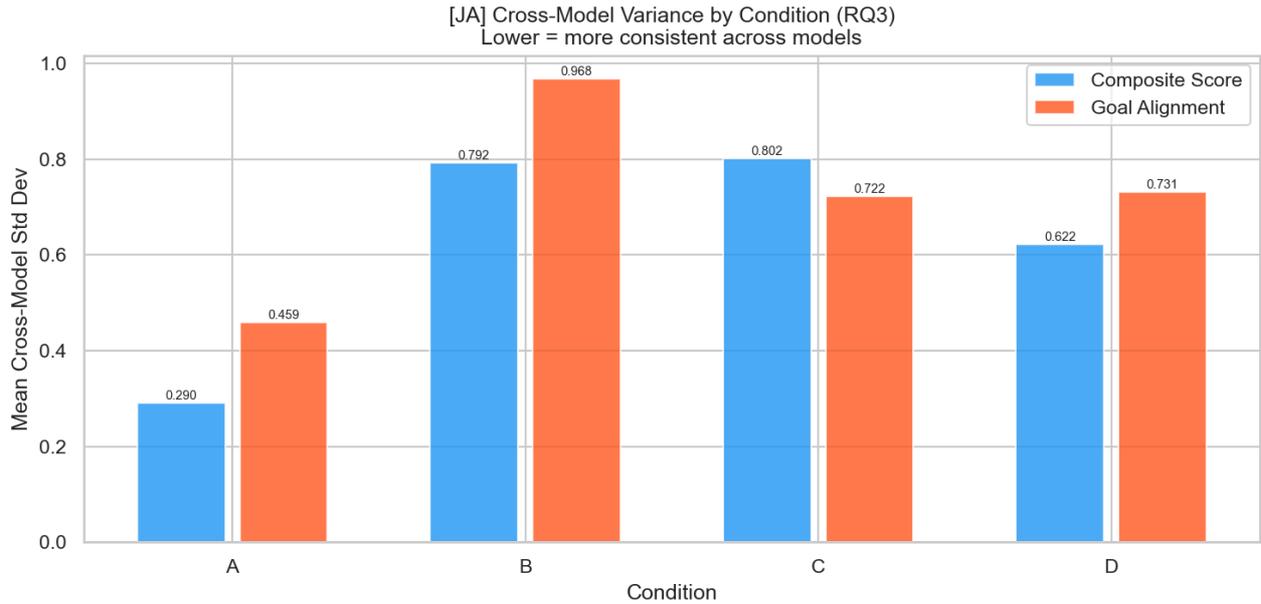

*Figure 2 (a–c). Mean cross-model standard deviation (σ) by condition. Lower σ indicates more consistent outputs across three LLMs. Blue: composite score; orange: goal alignment. Note: Condition A's low composite σ in ZH/JA is an inflation artifact (§5.3, §6.2).*

**Finding 3.1: A's low cross-model variance is an artifact, not a strength.**

In Chinese and Japanese, Condition A exhibits the lowest cross-model σ — superficially suggesting that unstructured prompts produce the most consistent cross-model outputs. However, this is an artifact of the same inflation mechanism identified in §5.1: when no constraints are specified, all three models tend to produce high-quality outputs on their own terms, yielding uniformly high scores and thus low variance. The consistency is spurious — it reflects models' unconstrained generative defaults converging on high scores, not shared alignment with a specified intent.

The goal alignment metric reveals this clearly: A's GA scores, while high in absolute terms, show higher cross-model variance than composite (ZH: σ_GA = 0.655 vs. σ_composite = 0.160). Different models infer different "most likely intents" from the same vague prompt, producing divergent outputs that all happen to score well on unconstrained quality metrics.

**Finding 3.2: Cross-model GA variance patterns are mixed but informative.**

On goal alignment — the intent-centered metric — the variance patterns across conditions are not uniform. Condition D achieves competitive cross-model variance compared to C in some but not all language-metric combinations (Table 4).

**Table 4. Cross-Model Standard Deviation — Goal Alignment**

| Condition | ZH σ (GA) | EN σ (GA) | JA σ (GA) |
|---|---|---|---|
| A | 0.655 | 0.481 | 0.459 |
| B | 0.788 | 0.504 | 0.968 |
| C | 0.454 | 0.496 | 0.722 |
| D | **0.551** | 0.625 | 0.731 |

The pattern is suggestive but not conclusive: in English, for instance, B shows lower GA σ than C or D (0.504 vs. 0.496/0.625), and in Japanese, C and D's GA σ values are similar to A's (0.722/0.731 vs. 0.459). The strongest consistency-related evidence comes not from structured conditions having uniformly low variance, but from demonstrating that Condition A's *apparent* low composite variance is an artifact of unconstrained generation (§6.2) — a finding that holds robustly across all three languages. The cross-model variance data are consistent with, but do not conclusively establish, the protocol interpretation of PPS.

# 6. Discussion

## 6.1 Democratizing Structured Prompting: C and D Show No Significant Difference

The absence of a statistically significant difference between Conditions C and D is the most practically significant finding of this study. Condition C represents expert-level prompt engineering: a user must understand the 5W3H framework, identify appropriate values for all eight dimensions, and express them in a structured format. This requires both domain knowledge and familiarity with the PPS schema.

Condition D reduces this to a single action: write one sentence describing your task. The AI-assisted authoring interface handles dimension expansion, and the user can optionally review and refine the result.

This creates a clear democratization pathway: - **Expert users** (Condition C): Full manual control over all dimensions - **Casual users** (Condition D): Single-sentence input with AI expansion - **Quality gap**: Statistically indistinguishable on goal alignment (the metric that matters)

Because Condition D in our experiment uses AI-generated expansions without user modification, it represents a *conservative lower bound*. In practice, users who review and adjust even one or two dimensions may achieve results closer to or better than Condition C — though this hypothesis requires validation through a controlled user study.

## 6.2 The Dual-Inflation Problem of Unstructured Baselines

Our results reveal a systematic bias in how unstructured prompts are evaluated that has broader implications for the field:

**Inflation 1 — Composite Score:** Condition A achieves the highest composite scores in all three languages. However, the *constraint_adherence* dimension — which rewards following explicit constraints — cannot penalize Condition A outputs, because Condition A specifies no explicit constraints. This creates a ceiling effect: A receives near-perfect constraint adherence scores by default, inflating its composite average.

**Inflation 2 — Cross-Model Variance:** In Chinese and Japanese, Condition A shows the lowest cross-model σ on composite scores. This appears to indicate high consistency, but as argued in §5.3, it reflects model defaults converging on high unconstrained scores — not genuine intent alignment.

Together, these inflations mean that using Condition A as a sole baseline will systematically underestimate the value of structured prompting. Evaluations of prompting approaches should: 1. Report both composite scores *and* goal alignment scores 2. Report cross-model variance alongside mean scores 3. Acknowledge that constraint-free baselines are not equivalent baselines

### 6.3 The Rendering Layer is Currently Essential

Condition B — which provides the same structured information as C and D in raw JSON format — consistently underperforms, particularly in English. This confirms the finding in [27] that the rendering layer is essential: current LLMs do not natively parse PPS-structured JSON as instruction intent.

This has an important implication for the B < A finding: it is *not* evidence that structured prompting is ineffective. It is evidence that structured information must be rendered into a form the model can process. The rendering layer bridges the gap between the structured representation (which captures intent in a more explicit and structured form) and the model's natural language processing capabilities.

As models improve their ability to process structured inputs, the rendering gap may narrow. However, for current-generation models, condition C and D consistently demonstrate that rendering matters.

### 6.4 The English Anomaly: Composite-GA Dissociation

English results exhibit a striking pattern that requires explanation: Condition C achieves high goal alignment (GA = 4.228) but low composite scores (2.899), while Condition A shows the reverse (composite = 4.126, GA = 4.489). This composite-GA dissociation is far more pronounced in English than in Chinese or Japanese.

We attribute this primarily to **format sensitivity in English-language LLMs.** The structured natural language rendering used in Conditions C and D — with labeled section headers ("Task Goal (What): ...", "Purpose (Why): ...") — appears to trigger a different generation mode in English than in Chinese or Japanese. English models may interpret the structured headers as a form template rather than substantive instructions, producing outputs that faithfully address the user's intent (high GA) but deviate from the expected document format or omit surface-level quality markers that the composite score rewards (lower structure, specificity, or overall_quality sub-scores).

This interpretation is supported by the domain breakdown data (Table B5): English Travel shows the most extreme dissociation (A composite: 3.293 vs. GA: 4.983), suggesting that open-ended English tasks are particularly susceptible to this format-content tension. The English results also show that D outperforms C on composite (3.482 vs. 2.899, *p* < 0.001) while being indistinguishable on GA — consistent with D's AI-expanded format being somewhat more natural-sounding than C's manually structured headers.

This finding has methodological implications: composite scores alone can be misleading when structured prompts change the surface format of outputs. Goal alignment, which evaluates intent satisfaction independently of format, provides a more robust comparison across conditions.

**6.5 PPS as a Structured Intent Layer: Evidence Summary**

Table 5 synthesizes the evidence for PPS's properties as a structured intent representation.

**Table 5. Properties of PPS — Evidence and Preliminary Observations**

| Property | Evidence |
| --- | --- |
| **Cross-language generalizability** | RQ2: Four consistent ordinal patterns hold across ZH/EN/JA |
| **Cross-model variance** | RQ3: Significant variance differences across conditions (Kruskal-Wallis *p* < 0.05 in all languages); A's low composite σ identified as inflation artifact |
| **Accessibility (zero-expertise path)** | No significant difference between C and D on GA: single-sentence input achieves results statistically indistinguishable from expert-level manual structuring |
| **Traceability** | SHA-256 hash + Instruction ID embedded in generated content *(design property; observed in preliminary deployment case, Appendix A)* |
| **Shareability** | Complete PPS specification can be shared and reused by collaborators *(design property; illustrated in Appendix A)* |
| **Cross-session persistence** | AI agent spontaneously reused PPS Instruction ID across dialogue rounds *(single-case observation, §6.6 and Appendix A; not tested in main experiment)* |

**6.6 Preliminary Deployment Observation**

We observed one real-world deployment case that illustrates PPS's accessibility and traceability properties outside the laboratory setting. A chemical researcher with no AI background used the AI-assisted authoring interface (Condition D pathway) to generate a PPS 5W3H instruction from a single sentence describing her research goal. The expanded instruction was submitted to MANUS (a multi-step AI agent), which produced a 14-page R&D proposal in a single pass. In a follow-up round, the researcher submitted an unstructured natural-language request — and the AI agent spontaneously reused the original PPS Instruction ID (`acf1387681f4`) in the second document's header, linking two independently generated outputs under a shared provenance record without explicit instruction to do so.

While this is a single case and cannot support generalizable conclusions, it illustrates two properties that are difficult to measure in controlled experiments: (1) the accessibility pathway from single-sentence input to professional-grade output via AI-assisted expansion, and (2) the potential for PPS metadata to function as a persistent task anchor across multi-round AI workflows. Full details of this case are provided in Appendix A.

### 6.7 Limitations

**Temperature confound between A and B/C/D.** Condition A uses temperature = 0.7 while Conditions B, C, and D use temperature = 0. This means that observed differences between A and structured conditions reflect the *joint effect* of structural completeness and sampling randomness — not structural completeness alone. In particular, Condition A's cross-model variance (RQ3) is influenced by both the absence of constraints and the stochastic generation setting. This confound does not affect the most important comparisons (C vs. D, C vs. B, D vs. B), which all share the same temperature setting, but it means that A-vs-C/D differences should be interpreted as upper bounds on the structural effect. A robustness experiment with all conditions at temperature = 0 would isolate the structural contribution more cleanly.

**Absence of an external gold-intent card.** The study does not compare all conditions against an external gold-intent specification defined independently of A/B/C/D. For Condition A, the judge infers the user's "most likely intent" from context; for Conditions C and D, the judge verifies alignment against the explicit 5W3H specification. This asymmetry means that part of C/D's advantage on goal alignment may reflect richer specification content available to the judge, rather than purely better transmission of a shared external intent. This limitation is more consequential in this paper than in Paper 1, because Condition D's AI-expanded specifications provide even more explicit detail for the judge to evaluate against.

**Model coverage.** All three generation models are Chinese-origin LLMs with strong multilingual capabilities. The generalizability of findings to GPT-4o, Claude, Gemini, and other model families remains to be tested.

**DeepSeek triple role.** DeepSeek-V3 (`deepseek-chat` API) serves three roles in this study: (1) Condition D expansion model, (2) one of three generation models, and (3) LLM-as-Judge. This overlap means DeepSeek evaluates outputs it may have generated (as generator) and may be predisposed toward the expansion style it uses (as expander). While §6.7's judge independence limitation addresses the generation-judge overlap, the expander-generator-judge triple role is a more specific confound: Condition D outputs expanded by DeepSeek may be evaluated more favorably by the same model serving as judge. Future work should use independent models for each role.

**User study scale.** The Intent-to-Use (ITU) survey from [27] (N=20) is not extended in this paper. A larger-scale user study specifically evaluating Condition D's usability and learning curve would strengthen the democratization argument.

**D expansion quality.** The quality of Condition D outputs is partially dependent on the quality of the AI expansion interface. A weaker expansion model would likely produce lower D scores.

### 6.8 Future Work

- **Extended model coverage:** Replication with GPT-4o, Claude, and Gemini to test cross-model generalizability beyond Chinese-origin LLMs

- **Temperature-controlled robustness study:** All conditions at temperature = 0 to isolate the structural contribution from sampling variability
- **External gold-intent evaluation:** Independent intent specification as ground truth for goal alignment assessment
- **User study:** Controlled experiment on Condition D usability, comparing novice vs. expert users, with pre/post user modification analysis
- **Cross-session persistence:** Systematic investigation of PPS ID reuse behavior across AI agents and session boundaries

## 7. Conclusion

This paper extends the empirical foundation of PPS (Prompt Protocol Specification) from a single-language study to a three-language, four-condition experiment covering 2,160 model outputs. Our principal findings are:

1. **AI-expanded 5W3H (Condition D) shows no statistically significant difference from manually crafted PPS (Condition C)** on goal alignment across Chinese, English, and Japanese — suggesting that AI-assisted authoring can democratize structured prompting without measurable loss in intent alignment.

2. **Four consistent patterns hold across all three languages:** C > B, D > B, no significant difference between C and D on goal alignment, and B ≪ A on composite score — providing strong evidence for PPS's cross-lingual generalizability.

3. **Structured conditions produce significantly different cross-model variance patterns** from unstructured baselines, with Condition A's apparent low variance revealed as an artifact of unconstrained generation rather than genuine consistency. The variance evidence is suggestive of protocol-like properties but is not uniform across all languages and metrics.

4. **Unstructured prompts exhibit dual inflation** — artificially high composite scores and artificially low cross-model variance — challenging their validity as isolated evaluation baselines.

Together, these findings suggest that structured 5W3H intent representations can serve as a useful human-AI interface layer — especially when combined with AI-assisted authoring that makes structured prompting accessible to non-expert users. The most practically significant result is the absence of a significant difference between C and D: a single-sentence input, when expanded by an AI authoring interface, achieves intent alignment statistically indistinguishable from expert-level manual structuring across three languages. Structured intent representations may not yet constitute a universal protocol, but they offer a concrete, empirically grounded step toward reducing the gap between what users intend and what AI models produce.

## References

[1] Tom B. Brown, Benjamin Mann, Nick Ryder, Melanie Subbiah, et al. Language models are few-shot learners. In *Advances in Neural Information Processing Systems (NeurIPS 2020)*, 2020. arXiv:2005.14165

[2] Jason Wei, Xuezhi Wang, Dale Schuurmans, Maarten Bosma, Brian Ichter, Fei Xia, Ed Chi, Quoc V. Le, and Denny Zhou. Chain-of-thought prompting elicits reasoning in large language models. In *NeurIPS 2022*, 2022. arXiv:2201.11903

[3] Shunyu Yao, Dian Yu, Jeffrey Zhao, Izhak Shafran, Thomas L. Griffiths, Yuan Cao, and Karthik Narasimhan. Tree of thoughts: Deliberate problem solving with large language models. In *NeurIPS 2023*, 2023. arXiv:2305.10601

[4] Pengfei Liu, Weizhe Yuan, Jinlan Fu, Zhengbao Jiang, Hiroaki Hayashi, and Graham Neubig. Pre-train, prompt, and predict: A systematic survey of prompting methods in natural language processing. *ACM Computing Surveys*, 55(9), Article 195, 2023. DOI:10.1145/3560815

[5] Sander Schulhoff, Michael Ilie, Noel Biest, et al. The prompt report: A systematic survey of prompting techniques. *arXiv preprint*, 2024. arXiv:2406.06608

[6] Laria Reynolds and Kyle McDonell. Prompt programming for large language models: Beyond the few-shot paradigm. In *Extended Abstracts of CHI 2021*, Article 341, pp. 1–7, 2021. DOI:10.1145/3411763.3451760

[7] Michael Xieyang Liu, Frederick Liu, Alexander J. Fiannaca, Terry Koo, Lucas Dixon, Michael Terry, and Carrie J. Cai. 'We need structured output': Towards user-centered constraints on large language model output. In *Extended Abstracts of CHI 2024*, 2024. DOI:10.1145/3613905.3650756 arXiv:2404.07362

[8] Saleema Amershi, Dan Weld, Mihaela Vorvoreanu, Adam Fourney, Besmira Nushi, Penny Collisson, Jina Suh, Shamsi Iqbal, Paul N. Bennett, Kori Inkpen, Jaime Teevan, Ruth Kikin-Gil, and Eric Horvitz. Guidelines for human-AI interaction. In *Proceedings of CHI 2019*, pp. 1–13, 2019. DOI:10.1145/3290605.3300233

[9] Ellen Jiang, Kristen Olson, Edwin Toh, Alejandra Molina, Aaron Michael Donsbach, Michael Terry, and Carrie Jun Cai. PromptMaker: Prompt-based prototyping with large language models. In *Extended Abstracts of CHI 2022*, 2022. DOI:10.1145/3491101.3503564

[10] Alireza Salemi, Sheshera Mysore, Michael Bendersky, and Hamed Zamani. LaMP: When large language models meet personalization. arXiv preprint, 2023. arXiv:2304.11406

[11] J.D. Zamfirescu-Pereira, Richmond Y. Wong, Bjoern Hartmann, and Qian Yang. Why Johnny can't prompt: How non-AI experts try (and fail) to design LLM prompts. In *Proceedings of CHI 2023*, Article 437, pp. 1–21, 2023. DOI:10.1145/3544548.3581388

[12] Lianmin Zheng, Wei-Lin Chiang, Ying Sheng, Siyuan Zhuang, Zhanghao Wu, Yonghao Zhuang, Zi Lin, Zhuohan Li, Dacheng Li, Eric P. Xing, Hao Zhang, Joseph E. Gonzalez, and Ion Stoica. Judging LLM-as-a-judge with MT-bench and chatbot arena. In *NeurIPS 2023 Datasets and Benchmarks Track*, 2023. arXiv:2306.05685

[13] Wei-Lin Chiang, Lianmin Zheng, Ying Sheng, Anastasios N. Angelopoulos, et al. Chatbot arena: An open platform for evaluating LLMs by human preference. In *Proceedings of ICML 2024*, PMLR 235:8359–8388, 2024. arXiv:2403.04132

[14] Sewon Min, Kalpesh Krishna, Xinxi Lyu, Mike Lewis, Wen-tau Yih, Pang Wei Koh, Mohit Iyyer, Luke Zettlemoyer, and Hannaneh Hajishirzi. FActScore: Fine-grained atomic evaluation of factual precision in long form text generation. In *Proceedings of EMNLP 2023*, pp. 12076–12100, 2023. arXiv:2305.14251

[15] Long Ouyang, Jeff Wu, Xu Jiang, Diogo Almeida, Carroll Wainwright, Pamela Mishkin, Chong Zhang, Sandhini Agarwal, Katarina Slama, Alex Ray, et al. Training language models to follow instructions with human feedback. In *NeurIPS 2022*, 2022. arXiv:2203.02155

[16] Vinton G. Cerf and Robert E. Kahn. A protocol for packet network intercommunication. *IEEE Transactions on Communications*, 22(5):637–648, 1974. DOI:10.1109/TCOM.1974.1092259

[17] Ian Arawjo, Chelse Swoopes, Priyan Vaithilingam, Martin Wattenberg, and Elena Glassman. ChainForge: A visual toolkit for prompt engineering and LLM hypothesis testing. In *Proceedings of CHI 2024*, 2024. DOI:10.1145/3613904.3642016 arXiv:2309.09128

[18] Yongchao Zhou, Andrei Ioan Muresanu, Ziwen Han, Keiran Paster, Silviu Pitis, Harris Chan, and Jimmy Ba. Large language models are human-level prompt engineers. In *Proceedings of ICLR 2023*, 2023. arXiv:2211.01910

[19] Swaroop Mishra, Daniel Khashabi, Chitta Baral, Yejin Choi, and Hannaneh Hajishirzi. Reframing instructional prompts to GPTk's language. In *Findings of ACL 2022*, pp. 589–612, 2022. arXiv:2109.07830

[20] Yizhong Wang, Yeganeh Kordi, Swaroop Mishra, Alisa Liu, Noah A. Smith, Daniel Khashabi, and Hannaneh Hajishirzi. Self-Instruct: Aligning language models with self-generated instructions. In *Proceedings of ACL 2023*, pp. 13484–13508, 2023. arXiv:2212.10560

[21] Q. Vera Liao and Jennifer Wortman Vaughan. AI transparency in the age of LLMs: A human-centered research roadmap. *Harvard Data Science Review*, Special Issue 5, 2023. arXiv:2306.01941

[22] Stephen H. Bach, Victor Sanh, Zheng-Xin Yong, Albert Webson, et al. PromptSource: An integrated development environment and repository for natural language prompts. In *Proceedings of ACL 2022: System Demonstrations*, pp. 93–104, 2022. arXiv:2202.01279

[23] Sunnie S. Y. Kim, Q. Vera Liao, Mihaela Vorvoreanu, Stephanie Ballard, and Jennifer Wortman Vaughan. 'I'm not sure, but...': Examining the impact of large language models' uncertainty expression on user reliance and trust. In *Proceedings of FAccT 2024*, 2024. DOI:10.1145/3630106.3658941 arXiv:2405.00623

[24] Murray Shanahan, Kyle McDonell, and Laria Reynolds. Role play with large language models. *Nature*, 623(7987):493–498, 2023. DOI:10.1038/s41586-023-06647-8

[25] Pranab Sahoo, Ayush Kumar Singh, Sriparna Saha, Vinija Jain, Samrat Mondal, and Aman Chadha. A systematic survey of prompt engineering in large language models: Techniques and applications. *arXiv preprint*, 2024. arXiv:2402.07927

[26] Gang Peng. *Super Prompt: 5W3H — A Comprehensive Guide to Designing Effective AI Prompts Across Domains*. Amazon KDP, April 2025. ASIN: B0F3Z25CHC.

[27] Gang Peng. Evaluating 5W3H Structured Prompting for Intent Alignment in Human-AI Interaction. *arXiv preprint*, 2026. arXiv:2603.18976 (March 2026)

# Appendix A: Real-World PPS Deployment Case

This appendix documents a real-world PPS deployment illustrating the framework's accessibility, traceability, and cross-session persistence properties (described in §6.7).

## A.1 User Context

A professional chemical researcher with no AI background used the lateni.com authoring interface to generate PPS instructions from a single-sentence task description. The researcher had prior laboratory experience with eutectic salt phase-change materials at 8°C and sought to extend into the −20°C to −5°C temperature range.

## A.2 PPS Instruction (English translation of original Chinese)

The following is the complete PPS 5W3H instruction generated by the AI-assisted authoring interface, reviewed and lightly edited by the researcher.

**PPS Standard:** v1.0.0 | **Instruction ID:** acf1387681f4 **Created:** 2026-03-25T09:13:15 **SHA-256:** 821e73daa5f8d02d506dd655850a05893940045bf3437be246b4acf1387681f4

**Task Goal (What):** Develop low-temperature inorganic phase-change cold storage materials with phase-change temperatures between −20°C and −5°C; prototype target: phase-change latent heat ≥ 200 kJ/kg; cycle stability > 1,000 cycles.

**Purpose (Why):** Address the urgent demand for efficient, stable cold storage technology in cold-chain logistics, medical refrigeration, and industrial cooling; leverage phase-change latent heat to improve energy utilization efficiency, reduce operational costs, reduce carbon emissions, and support sustainable development and green energy applications.

**Role & Audience (Who):** Cold-chain logistics companies (food and pharmaceutical transport), medical device manufacturers requiring low-temperature storage, industrial refrigeration system designers and engineers, research institutions in materials science and energy, government environmental agencies promoting energy-saving policies, and relevant industry associations developing standards.

**Timing (When):** 3-month R&D cycle including material screening, performance testing, and optimization phases; target initial prototype development within 3 months of project start; urgency driven by climate change policy mandates and accelerating market competition in cold-chain logistics.

**Context (Where):** Laboratory environment for material synthesis and performance characterization; application scenarios include global cold-chain networks, medical storage facilities, industrial refrigeration systems, and energy storage in cold-climate regions; geographic and regulatory considerations across North American, European, and Asian markets.

**Method (How to do):** Literature review and computational simulation to screen candidate inorganic materials (salt hydrates or eutectic salts) → Laboratory synthesis with doping modification to optimize phase-change temperature and thermal stability → Performance testing using DSC (Differential Scanning Calorimetry) and TGA (Thermogravimetric Analysis) → Pilot-scale validation for scalability and durability.

**Quantitative Requirements (How much):** Develop ≥ 2 high-performance material prototypes; phase-change latent heat ≥ 200 kJ/kg; cycle stability > 1,000 cycles.

**Expected Style (How feel):** Scientifically rigorous and innovative; professional and pragmatic tone emphasizing technical feasibility and application value; convey urgency and optimism to inspire team collaboration and market confidence; maintain objectivity to support reliable decision-making and long-term development.

### A.3 Round 1 Generated Output Summary

Submitted to MANUS (a multi-step AI agent), this instruction produced a 14-page R&D proposal in a single pass. The document header automatically embedded the PPS metadata:

> 指令ID: acf1387681f4 | PPS标准: v1.0.0 | 创建时间: 2026-03-25T09:13:15

**Document structure (10 sections):** 1. Project Overview (with 5W3H subsections) 2. Technical Background and Demand Analysis 3. Material Screening and Candidate Systems 4. Laboratory Synthesis and Modification Plan 5. Performance Testing and Characterization Protocols 6. Project Timeline and Milestones 7. Quantitative Targets and Acceptance Criteria 8. Risk Assessment and Mitigation Strategies 9. Application Scenarios and Market Analysis 10. References

**Selected content highlights:** - Thermal properties database for 12 eutectic salt candidate systems ($NaCl/H_2O$, $KCl/H_2O$, $NH_4Cl/H_2O$, $MgCl_2/H_2O$, etc.) with phase-change temperatures and latent heat values - Step-by-step lab synthesis protocol including nucleating agent screening (nano-$TiO_2$, $BaCO_3$, $SrCl_2$) and thickener modification (CMC-Na, sodium alginate) - DSC testing protocol with exact parameters (5°C/min, $N_2$ atmosphere, 50 mL/min flow) - 6-phase project timeline with milestone acceptance criteria - 4 peer-reviewed references (including 2025 publications)

The researcher reported this document was directly usable as a laboratory experiment guide.

### A.4 Round 2 Follow-Up Request (Unstructured)

The researcher submitted a natural-language follow-up — with no reference to the original PPS instruction:

> "前期我这边也调研过，就是用类似的共晶盐，现在我们完成的是8°，接下来要开发-20到-5°的低温相变材料。请你围绕这个目标进一步深挖关键技术，这次生成围绕关键技术的方案就可以，不用重复前面的内容"

(Translation: *"We have prior results with similar eutectic salts. Our 8°C system is now complete. Please deep-dive into the key technologies for the −20°C to −5°C target. Generate a focused technical plan — no need to repeat prior content."*)

**A.5 Round 2 Generated Output and Spontaneous PPS ID Reuse**

MANUS produced a second 14-page document focused on five key technical challenges. Critically, the document header **spontaneously reused the PPS Instruction ID without any explicit instruction:**

> *项目编号：PPS-acf1387681f4 | 版本：v1.0 编制日期：2026-03-25 | 适用阶段：实验室研发（承接8°C 体系已有成果）*

**Technical scope of Round 2 document (5 technical dimensions):** 1. **Supercooling control** — Classical Nucleation Theory (CNT) analysis; nucleating agent selection table with lattice-matching criteria (Telkes principle); T-history experimental protocol; target: supercooling ≤ 5°C 2. **Phase-change temperature control vs. latent heat retention** — Quantitative trade-off data (Yang et al. 2025: adding KCl to drop Tm from −15.6°C to −25.8°C causes 56% latent heat loss); recommended strategy: select materials with intrinsic Tm in target range 3. **Phase separation suppression for 1000-cycle stability** — CMC-Na + PAAS composite thickener design; accelerated cycling protocol (1000 cycles in ~1 month); surpasses current literature best (500 cycles at 76.5% retention) 4. **Thermal conductivity enhancement** — Graded strategy: nano-$Al_2O_3$ (0.25 wt%, +27%), SiC foam (3D matrix, +100–200%), CNT+$Al_2O_3$ composite (+50–80%); eliminates high-risk options (copper, carbon steel) 5. **Packaging material compatibility** — ASTM G31 corrosion protocol adapted for low-temperature; recommends HDPE; warns against PP embrittlement in the −10°C to −20°C range

The researcher confirmed this document was more immediately actionable ("更能落地") for laboratory implementation.

**A.6 Interpretation**

This two-round case illustrates that a PPS Instruction ID can function as a **persistent task anchor** across independent AI sessions — the agent recognized the ID from Round 1's context and voluntarily embedded it in Round 2's provenance header. The two documents together constitute a traceable R&D planning artifact: any future collaborator receiving either document can recover the original intent specification via the Instruction ID and verify content integrity via the SHA-256 hash. This is a single-case observation and cannot support generalizable conclusions about cross-session persistence; systematic investigation of this phenomenon remains future work.

## Appendix B: Statistical Tables

All pairwise comparisons use Mann-Whitney U tests (two-tailed). Cohen's *d* computed from rank-biserial correlation. Significance: * $p < 0.05$; ** $p < 0.01$.

**Table B1. Pairwise Comparisons on Composite Score and Goal Alignment — Chinese (ZH)**

| Comparison | Composite Δ | p | Cohen's d | GA Δ | p | Cohen's d |
|---|---|---|---|---|---|---|
| B vs. A | −0.838 | <0.001** | −1.556 | −0.250 | 0.004** | −0.297 |
| C vs. A | −0.311 | <0.001** | −0.820 | +0.261 | 0.013* | +0.373 |
| D vs. A | −0.513 | <0.001** | −0.939 | +0.150 | 0.055 | +0.183 |
| C vs. B | +0.527 | <0.001** | +0.829 | +0.511 | <0.001** | +0.712 |
| D vs. B | +0.324 | <0.001** | +0.434 | +0.400 | <0.001** | +0.478 |
| D vs. C | −0.202 | 0.050* | −0.315 | −0.111 | 0.812 | −0.160 |

$N = 180$ per condition. Bold: C ≈ D on GA ($p = 0.812$, not significant).

**Table B2. Pairwise Comparisons on Composite Score and Goal Alignment — English (EN)**

| Comparison | Composite Δ | p | Cohen's d | GA Δ | p | Cohen's d |
|---|---|---|---|---|---|---|
| B vs. A | −1.803 | <0.001** | −2.025 | −2.039 | <0.001** | −2.333 |
| C vs. A | −1.227 | <0.001** | −0.974 | −0.261 | <0.001** | −0.351 |
| D vs. A | −0.643 | <0.001** | −0.686 | −0.256 | 0.0002** | −0.324 |
| C vs. B | +0.577 | 0.001** | +0.554 | +1.778 | <0.001** | +2.127 |
| D vs. B | +1.160 | <0.001** | +1.889 | +1.783 | <0.001** | +2.036 |
| D vs. C | +0.583 | <0.001** | +0.539 | +0.006 | 0.633 | +0.007 |

*Note:* In English, D > C on composite ($p < 0.001$) but C ≈ D on GA ($p = 0.633$). English C composite is severely degraded by format mismatch; GA dissociates composite quality from intent alignment.

**Table B3. Pairwise Comparisons on Composite Score and Goal Alignment — Japanese (JA)**

| Comparison | Composite Δ | p | Cohen's d | GA Δ | p | Cohen's d |
|---|---|---|---|---|---|---|
| B vs. A | −1.530 | <0.001** | −2.420 | −1.244 | <0.001** | −1.488 |
| C vs. A | −0.529 | <0.001** | −0.818 | −0.200 | 0.027* | −0.281 |
| D vs. A | −0.628 | <0.001** | −1.018 | −0.250 | 0.017* | −0.343 |
| C vs. B | +1.001 | <0.001** | +1.186 | +1.044 | <0.001** | +1.109 |
| D vs. B | +0.902 | <0.001** | +1.099 | +0.994 | <0.001** | +1.042 |
| D vs. C | −0.099 | 0.108 | −0.119 | −0.050 | 0.772 | −0.059 |

*Note:* C ≈ D on both composite and GA in Japanese (both *p* > 0.05).

### Table B4. Cross-Model Standard Deviation by Condition and Language

Mean cross-model σ across 60 task pairs, for composite score and goal alignment. Bold = lowest σ per column.

| Condition | ZH σ (Composite) | ZH σ (GA) | EN σ (Composite) | EN σ (GA) | JA σ (Composite) | JA σ (GA) |
|---|---|---|---|---|---|---|
| A: Simple | **0.160** | 0.655 | 0.778 | 0.481 | **0.290** | 0.459 |
| B: Raw JSON | 0.743 | 0.788 | **0.464** | **0.504** | 0.792 | 0.968 |
| C: PPS NL | 0.452 | **0.454** | 0.806 | 0.496 | 0.802 | 0.722 |
| D: AI-Expanded | 0.571 | 0.551 | **0.587** | 0.625 | **0.622** | **0.731** |

Kruskal-Wallis H test across conditions:

| Language | Dimension | H statistic | *p* |
|---|---|---|---|
| ZH | Composite σ | 104.689 | <0.001** |
| ZH | GA σ | 25.082 | <0.001** |
| EN | Composite σ | 8.545 | 0.036* |
| EN | GA σ | 5.306 | 0.151 |
| JA | Composite σ | 67.771 | <0.001** |
| JA | GA σ | 36.008 | <0.001** |

### Table B5. Domain Breakdown — Mean Composite Score and Goal Alignment

*Chinese (ZH)*

| Domain | A Composite | A GA | B Composite | B GA | C Composite | C GA | D Composite | D GA |
|---|---|---|---|---|---|---|---|---|
| Business | 4.777 | 3.817 | 3.893 | 4.150 | 4.370 | 4.500 | 3.790 | 3.967 |
| Technical | 4.880 | 4.417 | 3.997 | 4.183 | 4.647 | 4.783 | 4.467 | 4.633 |
| Travel | 4.857 | 4.800 | 4.110 | 3.950 | 4.563 | 4.533 | 4.717 | 4.883 |

*English (EN)*

| Domain | A Composite | A GA | B Composite | B GA | C Composite | C GA | D Composite | D GA |
|---|---|---|---|---|---|---|---|---|
| Business | 4.357 | 3.883 | 2.463 | 2.833 | 2.440 | 3.733 | 3.233 | 4.050 |
| Technical | 4.727 | 4.600 | 2.430 | 2.117 | 3.790 | 4.583 | 3.687 | 4.117 |
| Travel | 3.293 | 4.983 | 2.073 | 2.400 | 2.467 | 4.367 | 3.527 | 4.533 |

*Note:* English Travel domain shows the largest A composite–GA dissociation (composite 3.293 vs. GA 4.983), reflecting A's artificial composite inflation when task is open-ended.

*Japanese (JA)*

| Domain | A Composite | A GA | B Composite | B GA | C Composite | C GA | D Composite | D GA |
|---|---|---|---|---|---|---|---|---|
| Business | 4.610 | 4.417 | 3.243 | 3.500 | 4.260 | 4.383 | 4.047 | 4.100 |
| Technical | 4.750 | 4.817 | 3.233 | 3.733 | 3.907 | 4.283 | 3.573 | 4.300 |
| Travel | 4.737 | 4.683 | 3.030 | 2.950 | 4.343 | 4.650 | 4.593 | 4.767 |

*Note:* In Japanese, D outperforms C on Travel both composite (4.593 vs. 4.343) and GA (4.767 vs. 4.650), suggesting that AI-expanded prompts may be particularly effective for open-ended, ambiguous tasks.